\documentclass[review]{elsarticle}

\usepackage{lineno,hyperref}
\usepackage{tabu}
\usepackage{graphicx}
\usepackage{geometry}
\usepackage{tabularx}
\usepackage{booktabs}
\usepackage{natbib}
\usepackage{array}
\usepackage{algorithm}
\usepackage{algorithmic}
\usepackage[center]{caption}
\usepackage{longtable}
\usepackage{rotating}
 \usepackage{amsmath}
\usepackage{subfigure}
\usepackage{pdflscape}
\usepackage{caption}
\usepackage{color}
\usepackage{graphicx}
\usepackage[OT1]{fontenc}

\modulolinenumbers[5]

\begin{document}

\begin{frontmatter}

\title{Military Dog Based Optimizer and its Application to Fake Review Detection}

\author{Ashish Kumar Tripathi}
\address{Jaypee Institute of Information Technology, Noida}
\address{mail2ashish07@gmail.com}
\author{Kapil Sharma}
\address{Delhi Technological University, New Delhi}
\address{kapil@ieee.org}
\author{Manju Bala}
\address{IP College of Women, New Delhi}
\address{manjugpm@gmail.com}

\begin{abstract}
Over the last three decades more then sixty meta-heuristic algorithms have been proposed by the various authors. Such algorithms are inspired from physical phenomena, animal behavior or evolutionary concepts. These algorithms have been widely used for solving the various real world optimization problems. Researchers are continuously working to improve the existing algorithms and also proposing new algorithms that are giving competitive results as compared to the existing algorithms present in the literature.
In this paper a novel meta heuristic algorithm based on military dogs squad is introduced. The proposed algorithm mimics the searching capability of the trained military dogs. Military dogs have strong smell senses by which they are able to search the suspicious objects like bombs, wildlife scats, currency, or blood  as well as  they can communicate with each other by their barking. The performance of the proposed algorithm is tested on 17 benchmark functions and compared with five other meta-heuristics namely particle swarm optimization (PSO), multi-verse optimizer (MVO), genetic algorithm (GA), probability based learning (PBIL) and evolutionary strategy (ES). The results are validated in terms of mean and standard deviation of the fitness value. The convergence behavior and consistency of the results have been also validated by plotting convergence graphs and BoxPlots. Further the, proposed algorithm is successfully  utilized to solve the real world fake review detection problem. The experimental results demonstrate that the proposed algorithm outperforms the other considered algorithms on the majority of performance parameters.
\end{abstract}


\end{frontmatter}

\section{Introduction} \label{sec:intro}

   {M}eta-heuristic algorithm are gaining more and more popularity in the engineering domain due to their ability to bypass local optima and applicability across different disciplines, whereas the classical optimization algorithms are not able to provide a suitable solution for solving the optimization problems of high dimensionality. Since, the search space increases exponentially with the problem size, therefore solving these problems with the techniques like exhaustive search is impractical. Various heuristic approaches have been developed by the researchers to solve the global optimization problems such as Genetic algorithm (GA) \cite{goldberg}, Particle swarm Optimization (PSO) \cite{kennedy2011}, Gravitational search algorithm (GSA) \cite{rashedi2009}, central force optimization (CFO) \cite{formato2007}, Colliding Bodies optimization (CBO) \cite{kaveh2014}, Magnetic charged system search (MCS) \cite{kaveh2014}, Ray optimization \cite{kaveh2013}, Cuckoo optimization (CO) \cite{rajabioun2011}, Firefly algorithm (FA) \cite{yang2010}, etc. Meta-heuristics are the population based algorithms inspired from the nature. Each algorithms starts with the random set of solution called population. What makes the difference is the way of movements of population towards the global optima during the optimization process. These algorithms are tested and analyzed in the different domains of engineering. As No Free Lunch theorem clearly obviates the claim of an optimization algorithm for all optimization problems \cite{wolpert}. Thus, the urge of new meta-heuristic algorithm is standstill. Therefore, in this paper, a new meta-heuristic algorithm is proposed which leverages the searching ability of the trained military dogs. Dogs are trained by the humans for object detecting and tracking purposes. They train them especially as military dogs, sniffer dogs, hunting dogs, police dogs, search dogs, and detector dogs. Military dogs are the category of dogs, especially trained for detecting substances like bombs, illegal drugs, wildlife scats, currency, or blood \cite{Detec87}. Mostly, military dogs work in groups called military dog squad to detect the object. They use the barking sound to locate or signal other dogs. Coren and Hodgson \cite{Under44} studied that each sound of the military dog have some meaning associated with it. For example, loud sound of dog indicates insecurity. Baying sound indicates a call from the military dog to assure that his mates are alerted \cite{Under44}.
Generally, the smelling power of the dog is 1,000 to 10,000 times more than the humans or other species \cite{Under44}. Table \ref{tab:scentReceptor} shows the number of scent receptors in the various species. Moreover, the military dogs have the capability of deducing the direction of smell by moving their nostrils. Also, they have ability of storing meaningful information about the object in the form of scent while searching, which helps them in reaching to the desired object.

\begin{table}
\scriptsize
\centering
\caption{Number of Scent Receptors for different Species}
\renewcommand{\arraystretch}{1.0}
\begin{tabular}{ l c}
\hline

\textbf{Species}	& \textbf{Number of Scent Receptors} \\

\hline
Humans	&	5 million	\\
Dachshund	&	125 million	\\
Fox Terrier	&	147 million	\\
Beagle	&	225 million	\\
German Shepherd	&	225 million	\\
Bloodhound	&	300 million	\\
    \hline
  \end{tabular}
\label{tab:scentReceptor}
\end{table}

Furthermore, united states war dogs association studies stated that the smelling power of dogs is effected by the wind. A dog may detect the suspected object up to 200 meters by the smell power if there is no wind. However, with the greater wind factor, the same can detect up to 1000 meters. Moreover, the factors like smoke and heavy vegetation are the confusing factors for a dog, as it confuse them in sensing the direction of actual smell or sound. This paper mimics the searching process of trained military dog squad to  introduce a novel military dog based optimizer for finding the global optima. 
The overall contribution of this paper has three folds. First, a new military dog based optimization has been presented. Second, the mathematical model of the proposed algorithm has been detailed. The validation of the proposed algorithm has been done against 17 benchmark functions and performance is measured in term of 4 parameters namely: fitness value, standard deviation, convergence behavior, and consistency in the results. The efficiency of the algorithm is compared with 5 existing meta-heuristics. Third, the real-world problem of fake review detection has been unfolded using the proposed algorithm.

Rest of the paper is organized as follows. Section II discusses the related work. Section III presents the mathematical model of the military dog based optimizer. Section IV provides the experimental results. Section V details the fake review detection problem and how it can be solved using MDBO. Conclusion and future work is elucidated in section VI.

\section{	Related Work}
Nature-inspired meta-heuristic algorithms mimics the optimization behavior of the nature. Generally, these algorithms are population-based and start with a population of random solutions to obtain the global best solution. In contrast to this, there exists single-solution based algorithms like hill climbing \cite{mitchell1993} and simulated annealing \cite{bandyopadhyay2008}, which initiates the optimization process with a single solution. However, these algorithms suffer with the problem of local trap and premature convergence as they do not share any kind of information. On the contrary, population-based algorithms improve the solution over the iterations by information sharing. 
Two common aspects of the population-based algorithms are exploration and exploitation. Exploration represents the diversification in the search space, while exploitation corresponds to the intensification of the current solution. All population based algorithm tries to attain an equilibrium between exploration and exploitation to achieve the global best solution. Every agent of the meta-heuristic tries to improve its performance by sharing its fitness value with other agents at each iteration. The meta-heuristic can be broadly classified into three categories namely; physics-based, swarm-behavior based and evolutionary-based. 

The physics-based algorithm optimizes the problem by imitating the physics based phenomenon.
Gravitational search algorithm, proposed by Rashedi et al. \cite{rashedi2009}, is one such algorithm which is based on Newtonian laws of gravity and  motion. Hosseini \cite{shah2009} proposed an intelligent water drop algorithm which was inspired from the flow of rivers, as rivers often follow shortest path while flowing from source to destination. Further, Birbil \cite{birbil2003} proposed an algorithm based on the concept of electromagnetism in which the properties of attraction and repulsion is used to attain a balanced trade-off between exploration and exploitation. Moreover, Mirjalili et al. \cite{mirjalili2016multi} proposed multi-verse optimizer (MVO) in 2015, which is based on the notion of cosmology i.e white hole, black hole and wormhole. Some other physics based algorithms are Galaxy-based Search Algorithm (GbSA) \cite{shah2011principal}, Black Hole (BH) \cite{hatamlou2013black} algorithm, Small-World Optimization Algorithm (SWOA) \cite{du2006small}, Ray Optimization (RO) \cite{kaveh2012new}, Curved Space Optimization (CSO) \cite{moghaddam2012curved}.\\

Swarm-based algorithms behave like the swarm of agents such as fishes or birds to achieve optimization results. Eberhart et al. \cite{optimization1995perth} proposed the particle swarm optimization (PSO) which was inspired from the swarming behavior of fish or birds in search of food. Gandomi \cite{gandomi2012} presented an algorithm based on the simulation of the krill individuals. Mirjalili \cite{mirjalili2015} proposed an ant-lion based optimizer that mimics the hunting mechanism of ant-lions. Moreover, Mirjalili \cite{mirjalili2015moth}also introduced the moth-flame optimization, which simulates the death behavior of moths, in which the movement of agent is based on the transverse orientation based navigation of moths. Further, Wang et al. \cite{wang2014hybrid} proposed the hybrid krill heard algorithm to overcome the problem of poor exploitation capability of the krill herd algorithm. Ant colony optimization is another swarm based algorithm, which imitates the path finding behavior of ants \cite{dorigo2011ant}. Some other swarm based algorithm proposed in the literature are Cuckko search, Bat algorithm, Firefly optimization, Spider monkey optimization and Artificial bee colony optimization \cite{karaboga2007powerful}.  \\
Evolution based algorithm are inspired from the biological evolution phenomena such as Darvin evolutionary theory. The evolutionary algorithms work on the principle of generating better individuals with the course of iterations by combining best individuals of the current generation. The popular genetic algorithm (GA) is an evolutionary algorithm based on the evolution of natural species. It maintains the balance between exploration and exploitation through the mutation and crossover operators. Another biological process based evolutionary algorithm is ES which gives almost equal importance to recombination and mutation, and it uses more than two parents to accord to an offspring. Baluja \cite{dasgupta2013evolutionary} proposed the probability-based incremental learning algorithm (PBIL) which manages only statics of the population rather than managing the complete population. Simon presented bio-geography based optimizer which is based on the immigration and emigration of the species between the islands of natural bio-geography. Differential evolution is another popular evolutionary algorithmic introduced by storm et al. \cite{storn1997differential}.

\section{Military Dog Optimizer}
In this section, a new optimization algorithm based on the behavior of military dog's squad is introduced.  
 Military dogs are the special trained dogs who go through a special training to search any specific type object, where they learn to identify thousands of scents. Moreover, military dogs undergo intense one on one training where they learn to work as a team to find the particular suspicious object for which they are trained.  
All these military dogs can communicate with each other by passing their message via barking. Hence dogs can cooperate with each other directly by passing message via the way of barking and its loudness. The loudness of barking indicates its closeness with the target object.
   When, a group of military dogs are left out for searching of a target object hidden in an open ground. The military dogs randomly start searching the area. With smelling sensation, the military dogs analyze a particular location and they define the fitness of the location in terms of loudness. The highest loudness indicates the best location among them.  
  Military dog takes small step based on the scent smell in a particular location to exploit the local search area and it moves to explore the search space based on the loudness of barking.
The smelling sensation analysis of the military dogs help them to take a move closer towards the target object and exploit the current location.  
  Military dogs diverge from each other to search for the target object and converge to indicate that the target object is close.
     \subsection{MDBO Definitions and Algorithm}
      In this subsection the behavior of military dogs is mathematically simulated and explained. First, some definitions for formalizing the MDBO are explained. Thereafter, the whole procedure of the MDBO is outlined. In the given definitions $R$ is used to refer the set of real numbers, $\phi$ is used to refer an empty set, while $Z$ is used to denote the set of integers.
    
  $Definition 1:$ A military dog squad $MDS^{m}$ is a set of $m$ trained military dogs. The size $m$ of the military dog squad remains constant. Future work could allow variable size military dog squad.\newline
    $Definition 2:$ The feasible solution vector $FSV^d$, represents the position of a military dog in $MDS$. $FSV \in R^d$ is a set of all real numbers that represents the urine marking of a $MDS$.  \newline
   $Definition 3:$  A military dog smell index $MDSI: MD \rightarrow R$ is a measure of goodness of the solution that is represented by a $MD$. In most of the population based algorithm, this $MDSI$ is called fitness of the individual.  \newline
 $Definition 4:$  Sniffing movement $\delta(p, P_m,):MD\rightarrow$ $MD$ is a probabilistic operator that randomly modifies the military dogs $FSV^d$ based on the fitness of the loudest barking $MD$ and movement probability $P_m$. Sniffing movement takes place using the following equation.
  \begin{equation}\label{eq:else5}
      FSV_i^j(t+1)=\begin{cases}FSV_{lodest}^{j}, \  p\leq P_m \\ FSV_i^j+R(0,1)\times step(i), \   p>P_m\end{cases}            
   \end{equation}
where,\newline
 ${  step(i)=w \times K(0, 1)\times ( FSV_i^j - FSV_{loudest}^j})$ \\
$K(0,1)$ is any randomly chosen number between 0 and 1. $w$ is wind constant and $p$ is any random number between 0,1.


  $Definition 5:$ Barking movement $\omega (p, q, \alpha):MD^n\rightarrow$ MD is a probabilistic operator that adjusts position of a military dog based on $FSV^d$ of loudest barking and any randomly chosen Military Dog. The probability $p$, that position of $MD$ is modified is constant and $q\in {(1,2,3...,d)}$ is the randomly chosen index. $alpha$ is the smog or vegetation constant that effects the sound coming from the other military dog. 

The feasible solution vector ($FSV^d$) modification of a military dog is defined by:
 \begin{equation}\label{eq:else5}
      FSV_i^j(t+1)=\begin{cases}FSV_i^j(t), \  p\leq \alpha \\ FSV_i^j(t)+B_m\times R(0,1), \   p>\alpha\end{cases}            
   \end{equation}

where $B_m= (FSV_{loudest}-FSV_{q})$ and $R(0,1)$ is any random number between (0,1).

    $Definition 6:$ The $MDS$ transition function $\phi =(m,d,\delta,\omega,P_m): MD^m\rightarrow MD^m$ is a 5-tuple that modifies the $MDS$ from one iteration to the next iteration. The $MDS$ transition function begins by computing the feasible solution vector $FSV^d$ and military dog smell index $MDSI$. Further, the $MDS$ modification is performed on each military dog $MD$ followed by $MDSI$ recalculation for each military dog. \newline                    
    $Definition 7:$ A MDBO algorithm $MDBO= (H,\phi,T)$ is a three tuple that finds the solution for an optimization problem. $H: \rightarrow \{MD^n , MDSI^n\}$ is a function that creates an initial $MDS$ and computes the corresponding $MDSI$. $\phi$ is a $MDS$ transition function defined earlier. $H$ is implemented using the random number generators inside the urine marking area of the military dog. $T :MD^n \rightarrow \{true,false\}$ is a termination criterion. \newline
 The MDBO algorithm can be informally described as follows:

          \begin{enumerate}
     \item The MDBO algorithm starts with the initialization of the MDBO parameters. In this step the method is derived for mapping the problem solution to  $FSV^d$ and $MDS$ as described in definition 1 and 2, which are problem dependent. Also the maximum number of military dogs, sniff movement probability $P_m$, smoke or vegetation constant $\alpha$, wind factor $K$ are initialized according to the nature of the optimization problem.   
\item Initialize the position of each military dog in the search space corresponding to the potential solution given in the problem. This is defined by the H operator described in definition 7.
\item Sniffing around current area (exploitation step):  In this step, each $MD$ modifies its $FSV$ based on the information got from the loudest barking dog.	 
While searching, the dogs take a random walk and steer around the new location. $MD$ searches around the target object and may either move directly towards the military dog at best position with movement probability $P_m$ or they may take random movements according to its own position and the position of the $MD$ nearest to the target object as described in definition 4. \newline 
   The pseducode of the sniffing movement is described as follows
             \begin{algorithm}
    \label{algo:sniffing movement}  
     \begin{algorithmic}
     \FOR  {$( i=1$ $ to$ $m )$} 
     \STATE IF{$(K<P_m)$}
     \STATE  $FSV_i^j(t+1)=FSV_{loudest}^{j}$
     \STATE   $step(i)=w \times K(0, 1)\times ( FSV_i^j - FSV_{loudest}^j);$
     \STATE $FSV_i^j{(t+1)}=FSV_i^j+R(0,1)\times step(i)$    
     \ENDFOR
     \end{algorithmic}
     \end{algorithm}
  
  \item Movement due to barking of other dogs(exploration step): It is the general nature of the military dogs that they bark loudly where they smell the suspected object. This creates a global movement of the military dogs. After a certain threshold of barking military dogs try to explore the search region with respect to the most loudly barking military dog. Each military dog takes a random move by considering the loudest barking military dog as global best and any randomly chosen barking military dog. The updated position is defined as per definition 5.
\begin{table*}
\scriptsize
  \caption{Parameter values of algorithm of proposed and other algorithms}
  \label{tab:ParaSett}
  \centering
  \begin{tabular}{p{0.1in} p{1.7in} l l l l l l l l l l}
   \toprule
    S. No. & Parameter & PSO & PBIL & GA & ES & MVO  & MDBO\\
    \midrule
    1.& Population Size ($N$) & 50 & 50& 50& 50& 50& 50 \\
  2.& Number of Iterations ($itr$) & 500 & 500 & 500 & 500 & 500 & 500 \\
  3.& Number of Dimensions ($dim$) & 30 & 30 & 30 & 30&30&30 \\
  4.&  Elite Size ($keep$)  & 2 & 2 & 2 & 2 & 2 & 2\\
  5.&  Inertial Constant ($w$) & $1$ & 0.3 & $--$ & $--$ & $--$ & $--$\\
  6.&  Congnitive Constant ($c_1$) & $1$ & -- & $--$ & $--$ & $--$ & $--$ \\
  7.&  Social Constant ($c_2$) & $1$ & -- & $--$ & $--$ & $--$ & $--$\\
  8.&  Mutation Probability ($P_{mutate}$)  & $--$ & $--$ & $.1$ & 0.1 \\
9.& Smog or Vegetation factor  ($\alpha$) & -- & --& --& --& --& .25\\
10.& Wind constant ($w$) & --& --& --& --& --&.25\\
    \bottomrule

  \end{tabular}
\end{table*}
               The psedo-code of the barking movement is described as follows:
     \begin{algorithm}
    \label{algo:barking movement}  
     \begin{algorithmic}
     \FOR  {$( i=1$ $ to$ $m )$}
     \STATE $K=rand(0,1)$ 
     \STATE K=IF{$(K<\alpha)$}
     \STATE    $ B_m=rand*(FSV_{loudest}-FSV_{q});$
     \STATE $FSV_i^{(i+1)}=FSV_i(t)+B_m\times K$    
     \ENDFOR
     \end{algorithmic}
     \end{algorithm}
       
\item Go to step three for the next iteration. This loop continues till the predefined number of iterations, or the desired solution has been found . This is the implementation of the T operator described in definition 6.
    \end{enumerate}
 Fig. \ref{fig:ee}a and Fig. \ref{fig:ee}b demonstrate the barking and sniffing movement of the $MDs$. It can be depicted from Fig. \ref{fig:ee}a that the barking movement represents the exploration step of the MDBO. However, the barking movement corresponds to the exploration step, as its movement is influenced any randomly chosen $MD$.	
 
\begin{table*}
\scriptsize
\caption{Benchmark Functions}
\renewcommand{\arraystretch}{1.8}
\centering
  \begin{tabular}{ p{0.1in}  p{0.7in}  p{2.5in}  p{0.43in}  p{0.23in}  p{0.63in} p{0.63in}}
    \hline
\textbf{Sr. No.}&  \textbf{Function Name}	&	 \textbf{Equation}		&		\textbf{Range }	 & \textbf{Optimal value } &\textbf{Optimal position values }&\textbf{Category}\\
\hline
1	&	Ackley	&	$ F_1(X)=-20  e^{-0.02 \sqrt{d^-1\sum_{i=1}^d x_i^2}}- e^{d^-1 \sum_{i=1}^d cos(2\pi x_i)} + 20 + e $	&	-32,+32	&	0	&	$(0,\cdots,0)$	&	Multi-Model	\\
2	&	Alpine	&	$F_2(X)=\sum_{i=1}^{d} |x_i \sin(x_i) + 0.1 x_i|$             	&	-100,+100	&	0	&	$(0,\cdots,0)$	&	Multi-Model	\\
3	&	Dixon and Price	&	$ F_3(X) = (x_1-1)^2 + \sum_{i=2}^{d}i(2x_i^2-x_{i-1})^2 $          	&	-100,+100	&	0	&	$(0,\cdots,0)$	&	Unimodal	\\
4	&	Griewank	&	$ F_4(X)=1 + \sum_{i=1}^{d} \frac{x_i^2}{4000} - \prod_{i=1}^d \cos(\frac{x_i}{\sqrt{i}}) $             	&	-20,+20	&	0	&	$(0,\cdots,0)$	&	Multi-model	\\
5	&	Levy	& $F_5(X)=\sin^2\left(\pi\omega_1\right)+\sum_{i=1}^{d-1}\left(\omega_i-1\right)^2\left[1+10\sin^2\left(\pi\omega_i+1\right)\right]+\left(\omega_d-1\right)^2\left[1+\sin^2\left(2\pi\omega_d\right)+\right], \omega_i=1+\frac{x_i-1}{4},for\space all\space i=1,\cdots,d$ &	-50,+50	&	0	&	$(0,\cdots,0)$	&	Multi-Model	\\
6	&	Pathological	&	$ F_6(X) = \sum_{i=1}^{d-1} \left( 0.5 + \frac{\sin^2 \sqrt{ 100x_i^2 + x_{i+1}^2} - 0.5} {1 + 0.001 (x_i^2 - 2x_ix_{i+1} + x_{i+1}^2)^2} \right) $	&	-100,+100	&	0	&	$(0,\cdots,0)$	&	Multi-model	\\
7	&	Perm	& $F_7(X)=\sum_{i=1}^{d} \left( \sum_{j=1}^{d}\left(j+\beta\right)\left(x_i^j-\frac{1} {j_i}\right) \right)^2$ &	-100,+100	&	0	&	$(1,1/2,\cdots,1/d)$	&	Multi-Model	\\
8	&	Powell            & $ F_8(X)$ = $ \sum_{i=1}^{d/4}$ $[  (x_{4i-3} + 10x_{4i-2})^2 + 5 (x_{4i-1}-x_{4i})^2 + (x_{4i-3} + 2x_{4i-2} )^4 + 10 (x_{4i-3} + x_{4i})^4]$   	&	-10,10	&	0	&	$(0,\cdots,0)$	&	Uni-model	\\
9	&	PowellSum          & $F_9(X)=\sum_{i=1}^{d} |x_i|^{i+1}$  	&	-100,+100	&	0	&	$(0,\cdots,0)$	&	Uni-Model	\\
10	&	Rastrigin	&	$ F_{10}(X) = 10d + \sum_{i=1}^d (x_i^2 - 10cos(2\pi x_i))$            	&	-5.12,+5.12	&	0	&	$(0,\cdots,0)$	&	Uni-Model	\\
11	&	Rosenbrock's	&	$ F_{11}(X) = \sum_{i=1}^{d-1} [100(x_{i+1} - x_i^2)^2 + (x_i-1)^2] $           	&	-30,+30	&	0	&	$(0,\cdots,0)$	&	Multi-Model	\\
12	&	Rotated Hyper-Ellipsoid	&	$ F_{12}(X) =	 \sum_{i=1}^{d} \sum_{j=1}^{i} x_j^2 $	&	-65.536,+65.536	&	0	&	$(0,\cdots,0)$	&	Uni-Model	\\
13	&	Schumer Steiglitz	&	$ F_{13}(X) = \sum_{i=1}^{d} x_i^4 $	&	-100,+100	&	0	&	$(0,\cdots,0)$	&	Uni-model	\\
14	&	Schwefel    	& $F_{14}=- \sum_{i=1}^d x_i \sin\sqrt{|x_i|} $	& -500,+500	&	0	&	$(0,\cdots,0)$	&	Multi-Model	\\
15	&	Sphere	&	$ F_{15}(X) =	\sum_{i=1}^{d} x_i^2 $	&	-100,+100	&	0	&	$(0,\cdots,0)$	&	Uni-Model	\\
16	&	Step	&	$ F_{16}(X) = \sum_{i=1}^{d} (\lfloor |x_i|\rfloor) $                 	&	-100,+100	&	0	&	$(0,\cdots,0)$	&	Uni-model	\\
17	&	Trigonometric	&	$ F_{17}(X) =\sum_{i=1}^{d} [d - \sum_{j=1}^d \cos x_j + i (1 - cos(x_i) - sin(x_i))]^2 $       	&	0,3.14	&	0	&	$(0,\cdots,0)$	&	Uni-Model	\\

    \hline
  \end{tabular}
\label{tab:BenchFn}
\end{table*}

\section{Fake Review Detection}
Nowadays, reviews play an important role in the sales of the products and services, thus ascertaining their authenticity is a challenging problem. For the same, fake review detection is one of the fundamental approach used to detect the fake reviews. In literature, the majority of the contemporary work is based on supervised learning models \cite{crawford2015survey}. However, the supervised models require labeled datasets. Therefore, the applicability of the supervised models is limited, as labeled datasets of fake reviews are rarely available. On the contrary, unsupervised learning models work on unlabeled datasets to induce the learning model. 
Generally, these models explore hidden structures of the dataset with $N$ data objects into $K$ clusters such that the data objects within a cluster have maximum resemblance \cite{kulhari2016unsupervised}. The traditional clustering methods, such as K-means and FCM, generally produce local optima in the presence of noise \cite{pandeyunsupervised}\cite{sharma2010selection} \cite{tripathidynamic} \cite{hatamlou2013black}. To alleviate this, the meta-heuristic algorithms have been proved to be efficient in performing clustering \cite{pandey2017twitter} \cite{ashish2018parallel}. Therefore, this paper leverages the strengths of MDBO to produce optimal cluster centroids for untangling the fake review detection problem.

 \subsection{MDBO based clustering for fake review detection}

In the MDBO based clustering, the $FSV$ of each military dog represents a set of cluster centroids, $C={\{C_1, C_2, \cdots ,C_k\}}$ for $K$ clusters. The $MDSI$ value of each military dog corresponds to the sum of squired Euclidean distance as defined in Eq.  (\ref{eq:25}).
   \begin{equation}\label{eq:25} 
    Min  D(Z,C)=\sum_{i=1}^N\sum_{j=1}^kw_{ij}\mid z_{i}-c_{j}\mid
   \end{equation}
Where $N$ represents the number of data objects, $\mid z_{i}-c_{j}\mid$ is the Ecludian distance of $I^{th}$ data object from the $J^{th}$ centroid. Further, $w_{ij}$ represents the association weight of $i^{th}$ review vector in the $j^{th}$ cluster, i.e. the value of $w_{ij}$ is 1 if the data object $i$ is allocated to the cluster $j$ otherwise 0. For, the $MDS$ of size $N$, the clustering process starts with $N$ candidate solution and these solutions are optimized with the course of iterations to improve $MDSI$. Finally, the $FSI$ of the $MD$ with best $MDSI$ value are returned as the final cluster centroids.

\subsection{Datasets}\label{sec:dataset}
 For performing the experiments, the real life dataset is collected from the Yelp \cite{Restaura5:online}, which has 142 million unique visitors from 31 countries. Yelp itself filters the reviews which are considered as highly reliable and accurate \cite{zhang2016online}.
  In this work, total of 6000 reviews are compiled from the Yelp recommended and non recommended section for genuine and fake reviews. The beautiful soup library of python for web scraping is used for crawling the data. The reviews are extracted pin code wise, starting from zip code 10000 and 10050, which corresponds to the restaurant pages of New York city. The complete block of the review is extracted consisting of review text, star rating by the reviewer to the restaurant, no of cool votes to the particular review, no of funny votes to the particular review, no of useful votes to the particular review, no of check-ins of reviewer in the hotel, no of photos uploaded by reviewer of the review on Yelp, no of friends of reviewer of the review on yelp, no of reviews till date of the reviewer of the review.
 Furthermore, the content analysis of the reviews has been performed using natural language tool kit (NLTK) to extract review centric features. NLTK provides an easy to use interface with rich set of lexical resources such as WordNet for the tokenization, parsing, semantic reasoning and tagging. Moreover, the feature extracted from the NLTK were also validated manually by randomly picking 30 reviews.  
  Total 11 verbal and non verbal features has been used in the experiments based on the previous studies \cite{rout2017deceptive} \cite{crawford2015survey} \cite{zhang2016online}. Table \ref{tab:fake-feature} contains the summary of the features used in the experiments. Each review represent a feature vector of length 11 and the numerical value of each feature is normalized between [0,1]. 

\begin{table*}
\scriptsize
\caption{Features taken for the clustering using MDBO}
\renewcommand{\arraystretch}{1.8}
\centering
  \begin{tabular}{ p{0.1in}  p{0.5in}  p{1.4in}  p{4.5in}  }
    \hline
\textbf{Sr. No.}&  \textbf{Feature Name}	&	 \textbf{Category}		&		\textbf{Definition}\\
\hline

1  &Nonverbal & Review count& It defines the total number of reviews posted by the reviewer.\\
2  & Nonverbal& check-ins: & it represents the number of check-ins by the reviewer for the hotel \\
3  & Nonverbal & Friend count:& It denotes the total number of friends of the person making review\\
4  & Nonverbal & Vote count & it represents the count of votes on the review\\
5 & Nonverbal & useful, cool, funny votes & count of useful, cool and funny votes \\
6  & Nonverbal & Followers & it is the total number of followers of the reviewers\\
7 & Nonverbal & Elite reviewer & it is the count of years for which reviewer has been a permanent yelp member\\
8 & Nonverbal & Average posting rate: & it represents the total number of reviews posted per day\\
9 & verbal & Review length & it is the total number of words per review\\
10 & verbal & Average content similarity & It is defines as the average similarity in the reviews given by a single reviewer\\
11 & Verbal & Average content similarity  & It corresponds the average similarity in the text of the reviews posted by a particular reviewer\\ 
 
    \hline
  \end{tabular}
\label{tab:fake-feature}
\end{table*}

\section{Experimental results}
  The performance of the proposed algorithm is evaluated in two folds, first the MDBO is validated on benchmark functions and results are are detailed in section \ref{sec:simulation result on benchmark functions}. Second, the the effectiveness of the MDBO is vindicated on fake review detection problem and the results are presented in section \ref{fake review det results}. For fair comparison, each algorithms is rum on a computer with 2.8 Ghz Intel ( R) Pentiam (R) core i3 processor and 8 GB RAM using Matlab 2015a.

\subsection{Benchmark Function Results}\label{sec:simulation result on benchmark functions}
  In this section, the performance and uniqueness of the proposed MDBO is analyzed and compared with five recent population based algorithms. Seventeen standard benchmark functions given in Table \ref{tab:BenchFn} are used for comparison of algorithm based on mean and standard deviation. Convergence behavior of MDBO is analyzed and compared with other algorithms by plotting the convergence graph for each benchmark function. Box plots are employed to visualize and establish the consistency of the proposed MDBO algorithm. Box plots are non parametric methods to display variations in results of proposed MDBO algorithm, which are further compared with five other algorithms on seventeen benchmark functions. Moreover, Wilcoxon rank sum test is performed, which shows the dissimilarity of MDBO with other algorithms.

\subsubsection{Comparison with existing algorithms}
 The proposed MDBO was tested on the minimization functions and the results were compared with five other algorithms namely MVO, ES, Pbil, GA as well as PSO. Table \ref{tab:ParaSett} contains the values of population size, number of dimensions, number of iterations, social constant, cognitive constant, mutation probability, wind factor and smog constant used in simulation. 
   Table \ref{tab:BenchFn} contains the details of the seventeen benchmark functions including range values, optimal position values and categories upon which the proposed algorithm has been tested and compared. Each function either belongs to uni-model or multi-model class. Nine uni-model functions are used to test the convergence rate and eight multi-model functions are used to test the local optima avoidance capability of the algorithm.
  Further, each algorithm was run fifteen times on each benchmark function to get the mean and standard deviation. Table \ref{tab:BenchmarkFunctionComparsion} shows the values of mean and the standard deviation of fitness values computed in fifteen rounds by each algorithm. From the comparison of mean and standard deviation of seventeen benchmark functions for six algorithms as given in Table \ref{tab:BenchmarkFunctionComparsion}, it is observed that proposed MDBO outperformed all five algorithms under comparison on sixteen benchmark functions in terms of mean fitness values. However, ES performed better than MDBO for only  one benchmark functions i.e F6 with mean value 4.48 as compared to 4.68 mean value of proposed MDBO. Further, standard deviation of the proposed MDBO is minimum for sixteen benchmark functions while ES has given minimum value of the standard deviation for one function i.e., F1. It can be observed that in all the nine uni-model functions proposed MDBO algorithm has beat all other algorithms showing stronger local search ability. However, proposed algorithm outperformed other algorithms in seven multi-model functions out of eight which confirms stronger exploration capability of the proposed algorithm.

\subsubsection{Wilcoxon Test}
      The uniqueness of the proposed algorithms have been statistically validated using Wilcoxon rank sum test. NULL hypothesis assumes that the two algorithms are similar at the five percent significance level $\alpha$ for benchmark functions. $p$ values has been computed for all the benchmark functions using the fitness values of compared and proposed algorithms. If the value of p$<$0.05 then null hypothesis is rejected and symbolized by `+' or `-', otherwise it is rejected and represented by symbol `='. However `+' indicates better result and `-'  represents poor results of the proposed MDBO algorithm. 
Table \ref{tab:wtest} shows the results of Wilcoxon rank sum test for the NULL hypothesis over seventeen benchmark functions explained Table \ref{tab:BenchFn}. The proposed MDBO algorithm is compared with ES, PSO, MVO, Pbil, and GA on the basis of $p$ values value. The $p$ value is computed by running fifteen iterations of each algorithm on all functions. A pair wise comparison of MDBO with other algorithms shows significant levels on the basis of $p$ value, mean and standard deviation. Significant level is positive if $p$ value is less than 0.05 and the value of mean and standard deviation are less than the compared algorithm. It is observed from the Table \ref{tab:wtest}, that MDBO has outperformed ES on all the benchmark functions except F6 where ES has given competitive result. Further, MDBO has surpassed PSO for all the benchmark functions. When MDBO is compared with MVO it has beaten on sixteen benchmark function out of seventeen. However for one function i.e., F6, GA performed well. Moreover, MDBO has given positive significance on all the benchmark functions when compared with PBIL and GA except for F6 function. 
  Hence is can be concluded that the proposed algorithm is significantly different and outperforms five existing algorithms i.e., MVO, ES, PBIL, GA as well as PSO on each benchmark function.


\begin{table*}
\tiny
\caption{Comparison of mean fitness and standard deviation values for 15 runs on benchmark functions for existing and proposed algorithms}
\renewcommand{\arraystretch}{1.0}

\begin{tabular}{p{0.10in} |p{0.35in}p{0.40in}| p{0.345in}p{0.365in}|p{0.345in}p{0.416in}|p{0.42in}p{0.41in}|p{0.35in}p{0.42in} |p{0.45in}p{0.42in}}
    \hline
  \textbf{Fun} & \multicolumn{2}{c}{\textbf{PBIL}} &			 \multicolumn{2}{c}{\textbf{PSO}}  	&	 \multicolumn{2}{c}{\textbf{GA}}   &		 \multicolumn{2}{c}{\textbf{MVO}}   &		 \multicolumn{2}{c}{\textbf{ES}} 	 & \multicolumn{2}{c}{\textbf{MDBO}}   \\

\cline{2-3}\cline{4-5}\cline{6-7}\cline{8-9}\cline{10-11}\cline{12-13}

    &		\textbf{Mean} &		\textbf{STD} &		 \textbf{Mean} &		\textbf{STD} &		 \textbf{Mean} &		\textbf{STD} &		 \textbf{Mean} &		 \textbf{STD} &	 \textbf{Mean} &		\textbf{STD} &		 \textbf{Mean} &		\textbf{STD}\\

\hline

$F_1$	&	4.51E+00	&	0.19E+00		&	3.55E+00	&	0.30E+00		&	2.18E+00	&	4.32E-01		&	1.63E+00	&	4.50E-01		&		6.31E+00		&	\textbf{1.88E-01}	&	\textbf{9.71E-01}	&	7.39E-01	\\
$F_2$	&	4.89E+02	&	3.54E+01		&	1.02E+00	&	5.01E+00		&	5.29E+01	&	1.15E+01		&	1.27E+02	&	3.42E+01		&		4.91E+02		&		3.67E+01			&	\textbf{2.34E+00}	&	\textbf{1.91E+00}	\\
$F_3$	&	8.56E+09	&	1.84E+09		&	3.06E+08	&	3.03E+08		&	1.05E+06	&	1.06E+06		&	5.71E+03	&	5.32E+03		&		8.26E+09		&		2.54E+09			&	\textbf{1.90E+02}	&	\textbf{7.28E+02}	\\
$F_4$	&	1.46E+00	&	0.06E+00		&	1.11E+00	&	0.04E+00		&	1.07E+00	&	2.57E-02		&	1.50E-02	&	1.25E-02		&		1.54E+00		&		4.98E-02			&	\textbf{2.79E-03}	&	\textbf{4.26E-03}	\\
$F_5$	&	3.69E+03	&	5.89E+02		&	5.21E+02	&	2.13E+02		&	5.32E+01	&	1.86E+01		&	4.83E+02	&	3.14E+02		&		3.70E+03		&		3.60E+02			&	\textbf{2.22E+01}	&	\textbf{2.35E+01}	\\
$F_{6}$	&	9.37E+00	&	0.66E+00		&	7.80E+00	&	0.52E+00		&	5.65E+00	&	4.94E-01		&	9.93E+00	&	3.62E-01		&	\textbf{4.48E+00}	&		\textbf{3.19E-01}			&		4.68E+00		&	8.03E-01	\\
$F_{7}$	&	2.00E+117	&	3.60E+117		&	3.30E+98	&	1.20E+99		&	5.59E+58	&	2.15E+59		&	1.05E+17	&	4.05E+17		&		4.50E+117		&		6.50E+117			&	\textbf{1.00E+10}	&	\textbf{0.00E+00}	\\
$F_{8}$	&	4.62E+04	&	93.54E+02		&	1.82E+04	&	7.40E+03		&	1.86E+02	&	1.22E+02		&	7.78E+00	&	5.49E+00		&		1.24E+05		&		2.91E+04			&	\textbf{9.65E-03}	&	\textbf{7.42E-03}	\\
$F_{9}$	&	4.46E+41	&	4.99E+41		&	1.32E+54	&	4.80E+54		&	3.33E+16	&	1.20E+17		&	2.96E+11	&	4.10E+11		&		8.41E+43		&		1.70E+44			&	\textbf{1.00E+10}	&	\textbf{0.00E+00}	\\
$F_{10}$	&	1.54E+02	&	1.17E+01		&	1.43E+00	&	3.94E+01	&		1.40E+01	&	6.49E+00		&	1.23E+02	&	3.56E+01		&		4.14E+02		&		1.96E+01			&	\textbf{2.07E+01}	&	\textbf{5.61E+00}	\\
$F_{11}$	&	1.39E+08	&	3.50E+07		&	3.67E+06	&	3.10E+06		&	9.47E+03	&	1.03E+04		&	4.58E+02	&	5.30E+02		&		1.50E+08		&		2.87E+07			&	\textbf{1.02E+02}	&	\textbf{4.02E+01}	\\
$F_{12}$	&	2.46E+01	&	1.99E+04		&	9.56E+04	&	7.77E+04		&	4.88E+03	&	2.50E+03		&	1.83E+01	&	1.31E+01		&		2.79E+05		&		2.70E+04			&	\textbf{8.89E-12}	&	\textbf{5.32E-12}	\\
$F_{13}$	&	1.87E+08	&	3.56E+08		&	7.40E+06	&	6.77E+06		&	4.25E+04	&	3.49E+04		&	1.01E-01	&	5.52E-02		&		1.55E+08		&		4.13E+07			&	\textbf{3.55E-16}	&	\textbf{5.74E-16}	\\
$F_{14}$	&	8.65E+00	&	3.32E+02		&	6.57E+03	&	1.07E+03		&	3.22E+03	&	6.44E+02		&	4.53E+03	&	7.70E+02	&		8.41E+03		&		3.79E+02			&	\textbf{5.48E+02	}	&	\textbf{2.20E+02}	\\
$F_{15}$	&	4.52E+00	&	4.14E+03		&	1.02E+04	&	2.70E+03		&	1.05E+03	&	4.98E+02		&	6.76E-01	&	1.91E-01		&		4.62E+04		&		4.21E+03			&	\textbf{2.07E-12}	&	\textbf{1.58E-12}	\\
$F_{16}$	&	9.32E+02	&	6.70E+00		&	4.82E+02	&	7.82E+01		&	1.82E+02	&	5.65E+01		&	5.60E+00	&	4.73E+00	&		9.12E+02		&		5.87E+01			&	\textbf{2.67E-01}	&	\textbf{7.04E-01}	\\
$F_{17}$	&	0.89E+00	&	2.36E+00		&	7.67E+03	&	6.86E+00		&	5.05E+00	&	1.96E+01		&	1.85E+00	&	9.18E-01		&		7.68E+03		&		2.36E+03			&	\textbf{2.74E-01}	&	\textbf{5.10E-01}	\\

  \hline
  \end{tabular}
\label{tab:BenchmarkFunctionComparsion}
\end{table*}



\begin{table*}
\caption{wilcoxon test for statistically significance level at $\alpha$ = 0.05 on benchmark functions}
\scriptsize
\begin{center}
\renewcommand{\arraystretch}{1.0}
  \begin{tabular}{c | p{0.45in} p{0.28in} |p{0.45in}  p{0.28in} |p{0.45in} p{0.28in}| p{0.45in} p{0.28in}| p{0.45in}  p{0.33in} }
    \hline
    \hline
\textbf{Function} & \multicolumn{2}{c}{\textbf{MDBO-ES}}  	&		 \multicolumn{2}{c}{\textbf{MDBO-PSO}}  &		 \multicolumn{2}{c}{\textbf{MDBO-MVO}}   &	 \multicolumn{2}{c}{\textbf{MDBO-PBIL}}   &	 \multicolumn{2}{c}{\textbf{MDBO-GA}} \\
\cline{2-3}\cline{4-5}\cline{6-7}\cline{8-9}\cline{10-11}
    &	$\textbf{p-value}$ &	$\textbf{SGFNT   }$ &	 $\textbf{p-value}$ &		 $\textbf{SGFNT }$ & 	 $\textbf{p-value}$ &	$\textbf{SGFNT }$ 	 &	 $\textbf{p-value}$  &		$\textbf{SGFNT }$ &		 $\textbf{p-value}$ &		$\textbf{SGFNT }$ \\
\hline

$F_1$ 	&	3.27E-06	&	 + 	&	3.27E-06	&	 + 	&	3.27E-06	&	 + 	&	3.27E-06	&	 + 	&	3.27E-06	&	 + 	\\
$F_2$ 	&	9.07E-06	&	 + 	&	3.39E-06	&	 + 	&	9.07E-06	&	 + 	&	9.07E-06	&	 + 	&	9.07E-06	&	 + 	\\
$F_3$ 	&	3.39E-06	&	 + 	&	3.39E-06	&	 + 	&	3.39E-06	&	 + 	&	3.39E-06	&	 + 	&	3.39E-06	&	 + 	\\
$F_4$ 	&	3.31E-06	&	 + 	&	3.31E-06	&	 + 	&	3.31E-06	&	 + 	&	3.31E-06	&	 + 	&	3.31E-06	&	 + 	\\
$F_5$ 	&	3.39E-06	&	 + 	&	3.39E-06	&	 + 	&	3.39E-06	&	 + 	&	3.39E-06	&	 + 	&	3.39E-06	&	 + 	\\
$F_6$ 	&	0.167962	&	 = 	&	3.39E-06	&	 + 	&	0.167962	&	 = 	&	0.167962	&	 = 	&	0.167962	&	 = 	\\
$F_7$ 	&	6.87E-07	&	 + 	&	6.87E-07	&	 + 	&	6.87E-07	&	 + 	&	6.87E-07	&	 + 	&	6.87E-07	&	 + 	\\
$F_8$ 	&	3.39E-06	&	 + 	&	3.39E-06	&	 + 	&	3.39E-06	&	 + 	&	3.39E-06	&	 + 	&	3.39E-06	&	 + 	\\
$F_9$ 	&	6.87E-07	&	 + 	&	6.87E-07	&	 + 	&	6.87E-07	&	 + 	&	6.87E-07	&	 + 	&	6.87E-07	&	 + 	\\
$F_{10}$ 	&	3.39E-06	&	 + 	&	3.39E-06	&	 + 	&	3.39E-06	&	 + 	&	3.39E-06	&	 + 	&	3.39E-06	&	 + 	\\
$F_{11}$ 	&	3.39E-06	&	 + 	&	3.39E-06	&	 + 	&	3.39E-06	&	 + 	&	3.39E-06	&	 + 	&	3.39E-06	&	 + 	\\
$F_{12}$ 	&	3.39E-06	&	 + 	&	3.39E-06	&	 + 	&	3.39E-06	&	 + 	&	3.39E-06	&	 + 	&	3.39E-06	&	 + 	\\
$F_{13}$ 	&	3.39E-06	&	 + 	&	3.39E-06	&	 + 	&	3.39E-06	&	 + 	&	3.39E-06	&	 + 	&	3.39E-06	&	 + 	\\
$F_{14}$ 	&	3.38E-06	&	 + 	&	3.38E-06	&	 + 	&	3.38E-06	&	 + 	&	3.38E-06	&	 + 	&	3.38E-06	&	 + 	\\
$F_{15}$ 	&	3.39E-06	&	 + 	&	3.39E-06	&	 + 	&	3.39E-06	&	 + 	&	3.39E-06	&	 + 	&	3.39E-06	&	 + 	\\
$F_{16}$ 	&	1.26E-06	&	 + 	&	1.26E-06	&	 + 	&	1.26E-06	&	 + 	&	1.26E-06	&	 + 	&	1.26E-06	&	 + 	\\
$F_{17}$ 	&	1.90E-06	&	 + 	&	1.90E-06	&	 + 	&	1.90E-06	&	 + 	&	1.90E-06	&	 + 	&	1.90E-06	&	 + 	\\

  \hline

  \end{tabular}
\end{center}
\label{tab:wtest}
\end{table*}

\subsubsection{Convergence rate}

The convergence behavior of the proposed MDBO algorithm is analyzed and compared with other five existing algorithms by plotting the convergence graph for each bench-mark function. Vertical axis of the graph represents the best of fitness value and the horizontal axis represents corresponding iteration number as depicted in Fig. \ref{fig:ConvergenceGraph1}. Further, Fig. \ref{fig:ConvergenceGraph1}a, \ref{fig:ConvergenceGraph1}b...\ref{fig:ConvergenceGraph1}q, shows convergence trends of six algorithms under study for benchmark functions F1 to F17 respectively. It can be visualized from the figures that the proposed MDBO is converging with a faster rate for fourteen benchmark functions out of seventeen benchmark functions as compared to MVO, ES, GA, PBIL and PSO. However for function F11, PSO and GA has outperformed MDBO and for function F13, F17 GA and ES are showing faster convergence rate respectively. It can also be concluded that the proposed algorithm is beating eighty eight percent of the uni-model functions and seventy eight percent of the multi-model functions. The results shows that the proposed algorithm not only gives better fitness value but also shows good convergence rate for both the unimodal and multi-modal functions.

\subsubsection{Box plots}
 Box plots are used to show the consistency in the final result values found for each compared and proposed algorithm over fifteen runs. Fig. \ref{fig:Boxplot1} depicts groups of final best solution values through their quartiles. Further extending vertical lines from the boxes indicates variability outside the upper and lower quartiles of final best solution for all algorithms under comparison. Fig. \ref{fig:Boxplot1} contains seventeen sub-figures a, b, c...q for seventeen benchmark functions  F1, F2...F17 respectively as given in Table \ref{tab:BenchFn}. Fig. \ref{fig:Boxplot1}d, \ref{fig:Boxplot1}e, \ref{fig:Boxplot1}j and \ref{fig:Boxplot1}p shows minimum spacing between the different parts of boxes as compared to ES, PSO, MVO, PBIL and GA that indicates the degree of dispersion and skewness in the final best solution. And hence the proposed MDBO has outperformed ES, PSO, MVO, PBIL and GA algorithms for functions F4, F5, F10 and F16. The proposed MDBO has beat ES, PSO and PBIL and tie with MVO and GA as observed from Fig. \ref{fig:Boxplot1}c, \ref{fig:Boxplot1}h, \ref{fig:Boxplot1}k, \ref{fig:Boxplot1}m. Fig. \ref{fig:Boxplot1}i shows a tie for all algorithms under study. Further \ref{fig:Boxplot1}g and \ref{fig:Boxplot1}n depicts that MDBO beat ES and PBIL and tie with PSO, MVO and GA. Fig. \ref{fig:Boxplot1}b depicts that MDBO beat ES, PSO, PBIL and GA and tie MVO. Fig. \ref{fig:Boxplot1}q depicts MDBO beat ES, PSO and tie with MVO, PBIL and GA. The proposed MDBO is defeated by all algorithms under study as depicted in Fig. \ref{fig:Boxplot1}a and \ref{fig:Boxplot1}f. Further MDBO beat PSO, MVO, PBIL and GA except ES as shown in Fig. \ref{fig:Boxplot1}e. 

\subsection{Fake Review Detection Results}\label{fake review det results}
 This section, details the performance of the MDBO for fake review detection problem. The dataset used for evaluating the performance is explained in section \ref{sec:dataset}. The proposed algorithm has been validated by comparing the results with K-means and 4 other meta-heuristic based algorithms proposed for the clustering. Table \ref{tab:frr} presents the mean and best accuracy of the proposed and considered algorithms obtained by executing each algorithm 15 times in the same environment.
 It can be depicted from the table \ref{tab:frr}, that the proposed method has outperformed the K-means and other considered meta-heuristic based clustering methods for the fake review detection. Thus, it can be concluded that the proposed method can serve as a powerful tool for solving the fake review detection problem. 
\begin{table*}
\caption{Best and average accuracy over 15 runs}
\begin{center}
\renewcommand{\arraystretch}{1}
  \begin{tabular}{l l l l l l l r}
    
    \hline
\textbf{Dataset Name} & \textbf{Criteria} 		&		\textbf{K-Means}     &	 \textbf{PSO}  &\textbf{GSA} &\textbf{GA}&\textbf{GWO}&\textbf{MDBO} \\
\hline
Yelp review	&Best & 50.24 &54.32 &57.60 &53.46 & 62.32 &\textbf{71.42}\\
                    & Mean & 50.72 &58.46 &54.68 &59.50 & 63.92 &\textbf{71.52}	\\

\hline
  \end{tabular}
\end{center}
\label{tab:frr}
\end{table*}

\section{Conclusion}
  In last three decades more than sixty meta-heuristic algorithms have been proposed by the various authors. Some of them are inspired by physics based phenomena of nature like GSA, CFO and Black hole optimization and some are inspired by the swarm behavior and evolution of species. This paper proposed a new optimization algorithm called Military Dog Based Optimization (MDBO). MDBO utilizes the searching capability of the suspected objects by the trained military dogs. MDBO uses the tendency of military dogs to communicate with each other and move towards the suspected object by their smell power. Every Military Dog follow his companion which have the best fitness value. 
 The overall contribution of the paper has been divided into four folds, (i) a novel algorithm has been proposed (ii) The proposed algorithm has been validated on 17 benchmark functions and compared with 5 other existing meta-heuristics(iii) The proposed algorithm is successfully utilized for the fake review detection problem. The proposed MDBO outperforms the other algorithms in terms of mean fitness, standard deviations and convergence behavior. The consistency in the results of the MDBO has been statistically validated using the Box-plots curves. The quality of the results and the uniqueness of the MDBO has been also verified by the Wilcoxon rank sum test. Further, the proposed algorithm has also surpassed the compared algorithm in detecting the fake reviews. Thus, it can be concluded that the proposed algorithm can be applied for the optimization problem and can also serve as an alternative tool for handling real word problems. The future work will include the applications of the proposed algorithm in various real world large datasets optimization problems. Moreover, it can also be tested for the big data problems by adopting it in parallel environment.

\end{document}